\documentclass[11pt]{article}

\usepackage[final]{acl}

\usepackage{times}
\usepackage{latexsym}

\usepackage[T1]{fontenc}

\usepackage[utf8]{inputenc}

\usepackage{microtype}

\usepackage{inconsolata}

\usepackage{graphicx}
\usepackage{booktabs}
\usepackage{algorithmicx}
\usepackage{algorithm}
\usepackage{algpseudocode}
\usepackage{amsmath}
\usepackage{multirow}
\usepackage{multicol}
\usepackage{enumitem}
\usepackage{tabularx}

\title{CodeDelegator: Mitigating Context Pollution via Role Separation in Code-as-Action Agents}

\author{
	Tianxiang Fei\footnotemark[1],
	Cheng Chen\footnotemark[1],
	Yue Pan\footnotemark[1],
	Mao Zheng,
    Mingyang Song
	\\
	\\Large Language Model Department, Tencent\\
	{ \{alvinfei, cianchen, nobitapan, moonzheng, nickmysong\}@tencent.com}  \\
}

\begin{document}
\maketitle
\renewcommand{\thefootnote}{\fnsymbol{footnote}}
\footnotetext[1]{These authors contributed equally to this work.}
\renewcommand{\thefootnote}{\arabic{footnote}}

\begin{abstract}
Recent advances in large language models (LLMs) allow agents to represent actions as executable code, offering greater expressivity than traditional tool-calling. However, real-world tasks often demand both strategic planning and detailed implementation. Using a single agent for both leads to context pollution from debugging traces and intermediate failures, impairing long-horizon performance. We propose \textbf{\textsc{CodeDelegator}}, a multi-agent framework that separates planning from implementation via role specialization. A persistent \textit{Delegator} maintains strategic oversight by decomposing tasks, writing specifications, and monitoring progress without executing code. For each sub-task, a new \textit{Coder} agent is instantiated with a clean context containing only its specification, shielding it from prior failures. To coordinate between agents, we introduce \textbf{Ephemeral-Persistent State Separation (EPSS)}, which isolates each Coder’s execution state while preserving global coherence, preventing debugging traces from polluting the Delegator’s context. Experiments on various benchmarks demonstrate the effectiveness of \textsc{CodeDelegator} across diverse scenarios.
\end{abstract}

\section{Introduction}

\begin{figure}[ht]
\centering
\includegraphics[scale=0.128]{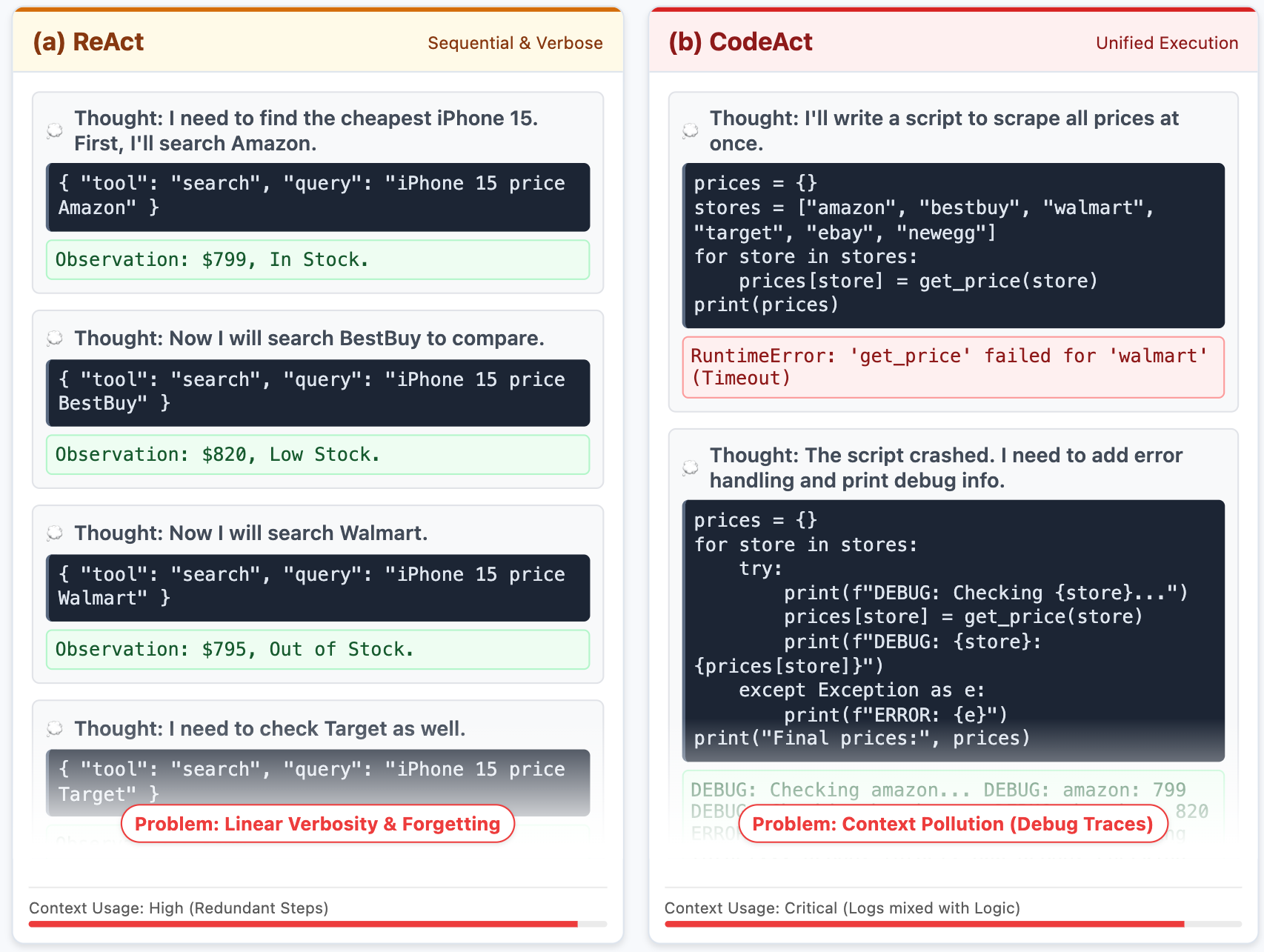}
\caption{Comparison of action representations: ReAct uses structured text/JSON, while CodeAct uses executable Python code, enabling more complex logic within a single action}
\label{fig: fig_intro_1}
\end{figure}

The integration of external tools into Large Language Models (LLMs) marks a fundamental advance, enabling complex problem-solving through real-world interaction~\cite{qu2025tool, liu2025advances}. ReAct~\cite{yao2023react} exemplifies this capability through an iterative \textit{Thought-Action-Observation} loop. However, ReAct-based frameworks typically represent actions as JSON or structured text, which constrains their expressiveness to predefined schemas and leads to linearly increasing context length, thereby reducing execution efficiency (Fig.~\ref{fig: fig_intro_1}(a)). The ``code-as-action'' paradigm (e.g., CodeAct~\cite{wang2024executable}) addresses this by adopting executable Python as a unified action representation, naturally supporting control flow, variable management, and multi-tool composition within a single step.

\begin{figure*}[ht]
\centering
\includegraphics[scale=0.25]{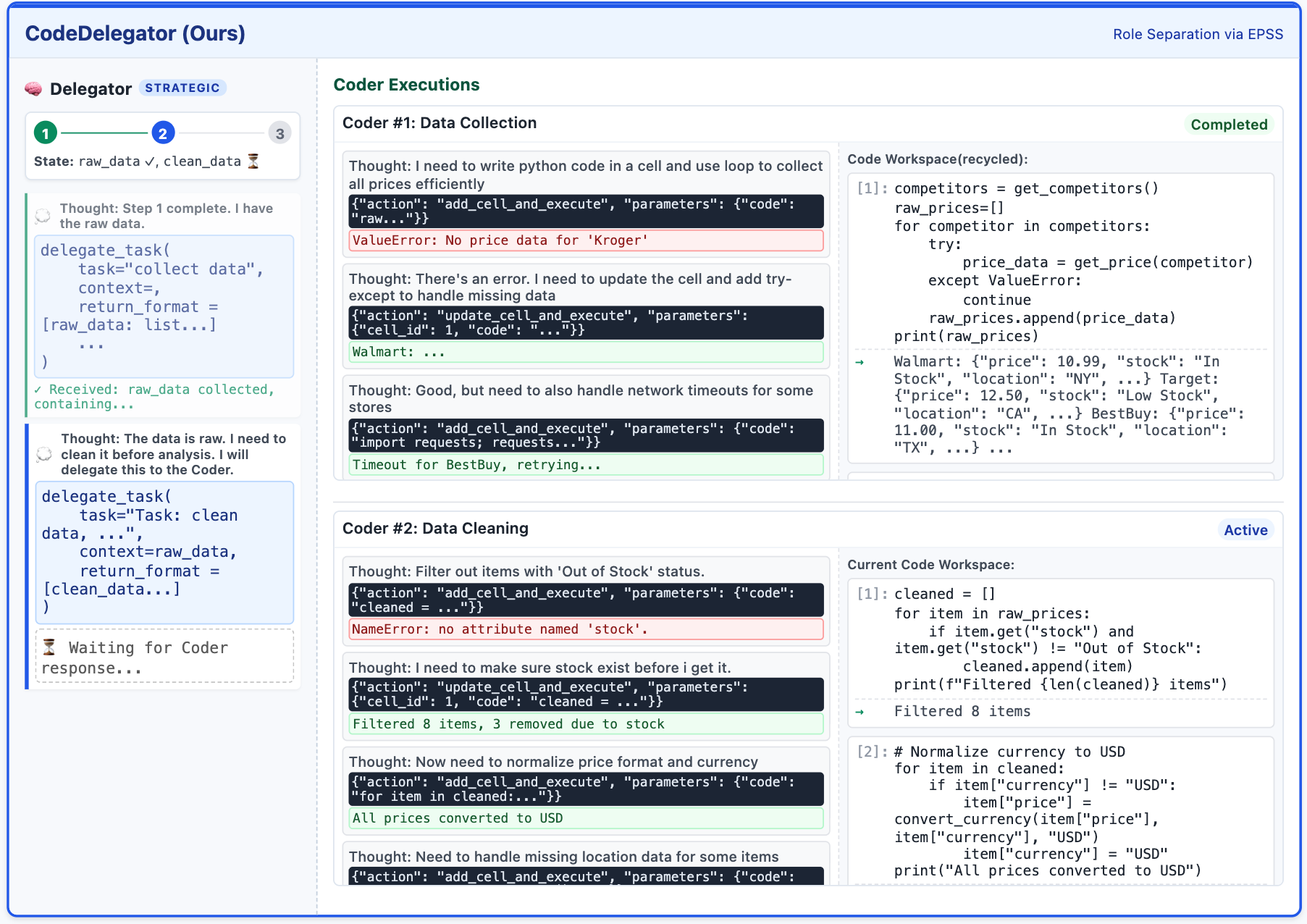}
\caption{CodeDelegator in action: The persistent Delegator orchestrates 
a pricing analysis task by delegating sub-tasks to ephemeral Coders. 
Coder \#1 (Data Collection) has completed and been discarded; Coder \#2 
(Data Cleaning) is actively debugging in an isolated context. Each Coder's 
execution traces (errors, retries) remain confined and never propagate 
to the Delegator's planning context, preventing context pollution.}
\label{fig: fig_intro_2}
\end{figure*}

However, when handling real-world, long-horizon tasks that interact with complex environments and humans, code-as-action agents encounter a fundamental tension. Consider a task ``analyze competitor iPhone 15 pricing data and generate a market report'': the agent must strategically decompose this into sub-tasks (data collection, data cleaning, analysis report generation) and reason about dependencies. Simultaneously, for each sub-task, it must generate syntactically correct code with proper API calls and error handling. These objectives compete for attention within a fixed context window, where later tokens progressively overshadow earlier content~\cite{liu-etal-2024-lost}. Moreover, planning requires abstract reasoning, while implementation demands fine-grained attention to detail. When interleaved, planning steps and debugging traces accumulate in shared context. We refer to this as \textit{context pollution}: the progressive dilution of task-relevant information that degrades both planning and implementation capabilities. (Fig.~\ref{fig: fig_intro_1}(b)).

To address this tension, we propose \textbf{\textsc{CodeDelegator}}, which decouples strategic planning from code implementation through role separation (Fig.~\ref{fig: fig_intro_2}). A Delegator maintains strategic focus: decomposing complex tasks, analyzing dependencies, monitoring progress, and dynamically dispatching work, without ever writing implementation code. Ephemeral Coders execute assigned sub-tasks through an interactive loop: generating code, observing execution results, diagnosing errors, and iteratively refining until completion. Each Coder is spawned by the Delegator with a fresh context containing only task-relevant information, preventing accumulation of irrelevant history from other sub-tasks.
A key challenge is enabling effective coordination without reintroducing context pollution. We address this through \textbf{Ephemeral-Persistent State Separation (EPSS)}, a dual-layer workspace providing code-aware isolation beyond mere conversation separation. The Delegator governs a persistent Orchestration Layer containing global state, task plans, and committed results. Each Coder operates in an ephemeral execution sandbox where debugging traces and local variables remain confined and are discarded upon completion. Structured schemas govern all inter-agent communication, replacing lossy natural language summaries with typed specifications.
Our contributions are:
\begin{itemize}[topsep=0pt, partopsep=0pt, itemsep=1pt, parsep=0pt]
\item We identify and empirically characterize the {context pollution} problem in code-as-action agents. (\S~\ref{sec:analysis}).
\item We propose \textsc{CodeDelegator} (\S~\ref{sec:methodology}), a two-role architecture that achieves role separation, complemented by EPSS for effective coordination.
\item Experiments (\S~\ref{sec:exp}) on $\tau^2$-bench and MCPMark demonstrate that \textsc{CodeDelegator} achieves substantial improvement over baselines.
\end{itemize}

\section{Related Work}

The remarkable reasoning capabilities of Large Language Models have spurred research into autonomous agents that use tools to solve complex tasks~\cite{wei2022chain, huang2022inner, yao2023react, qu2025tool}. The \textit{plan-execute-observe} loop introduced by ReAct~\cite{yao2023react} has become foundational for many agent systems~\cite{yang2023auto, wu2024autogen, zhang2024aflow, hu2025owl}.

\subsection{Code-as-Action LLM Agents}

CodeAct~\cite{wang2024executable} demonstrates that executable Python code naturally supports control flow, variable management, and multi-tool composition, outperforming ReAct on complex tasks. This paradigm has been adopted by OpenHands~\cite{wang2025openhandssoftwareagentsdk}, SWE-agent~\cite{yang2024sweagentagentcomputerinterfacesenable}, and Voyager~\cite{wang2024voyager}. 
Recent work explores hierarchical code generation to enhance planning ability. \citet{liu2025interactive} generates abstract code skeletons for progressive refinement. \citet{yu2025recode} introduces recursive decomposition of placeholder functions. These approaches address \textit{what} to implement versus \textit{how}. However, planning and implementation still share a single context, so debugging traces from implementing step $i$ accumulate when planning step $i{+}1$. \textsc{CodeDelegator} addresses an orthogonal \textit{isolation} concern: the Delegator and Coders operate in physically separate contexts, ensuring that no implementation detail, regardless of abstraction level, ever reaches the planning process.

\subsection{Multi-Agent Collaboration}

Multi-agent collaboration leverages specialized roles for complex tasks, as exemplified by AutoGen~\citep{wu2024autogen}, MetaGPT~\citep{hong2023metagpt}, and CAMEL~\citep{li2023camel}. However, \citet{cemri2025multi} identified recurring coordination failures including specification drift, inter-agent misalignment, and verification gaps.
\textsc{CodeDelegator} differs from general multi-agent frameworks in two key aspects: (1) agents are ephemeral rather than persistent, preventing cross-task context pollution; (2) information flow is structurally asymmetric, with execution traces confined to worker agents and never propagated to the planner. These designs specifically target the context pollution problem in code-as-action settings.

\begin{figure*}[!ht]
    \centering
    \includegraphics[scale=0.25]{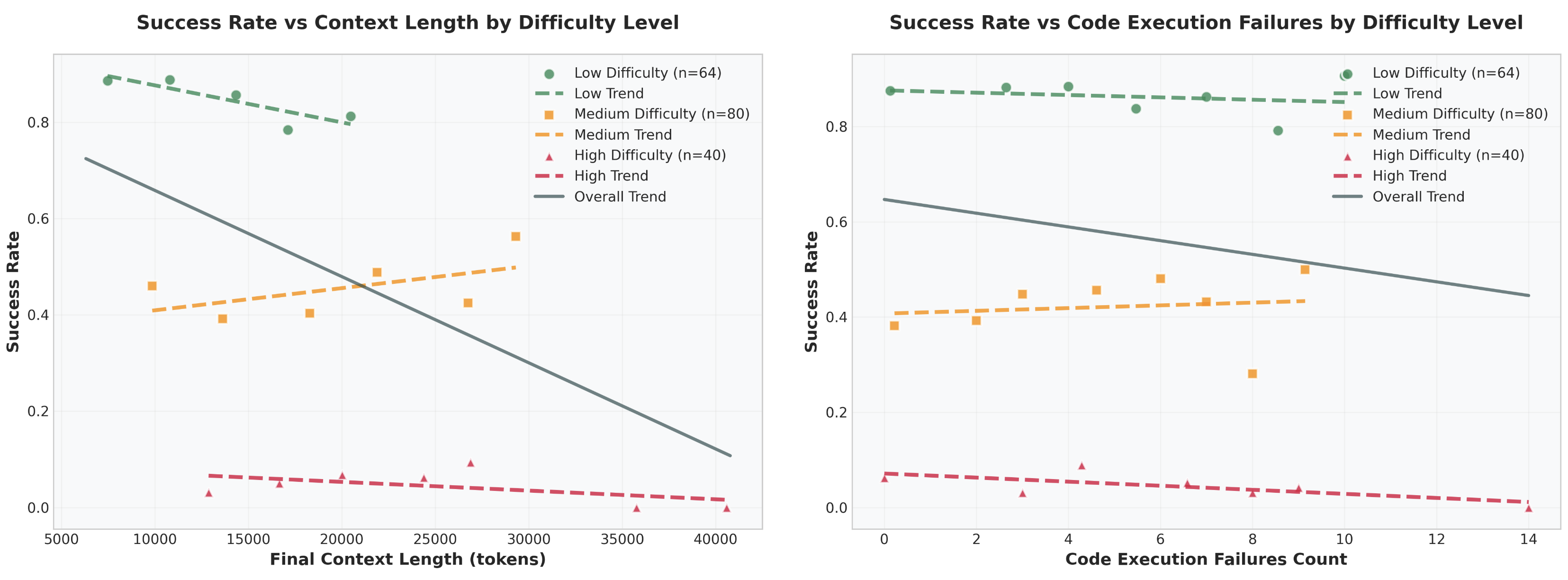}
    \caption{Illustrative examples of agent trajectories and failure modes observed in the pilot study.}
    \label{fig:pre_analysis}
\end{figure*}
\section{Understanding Context Pollution in Code-as-Action Agents}
\label{sec:analysis}

We define \textit{context pollution} in code-as-action agents as the progressive dilution of task-relevant information in shared context: debugging traces from sub-task $s_i$ persist during $s_j$, competing for attention when they are irrelevant. To validate that pollution, not task 
difficulty alone, limits code-as-action agents, we conduct a pilot study on CodeAct.

\paragraph{Setup}
We evaluate CodeAct on 50 tasks from the airline domain of $\tau^2$-Bench~\cite{barres2025tau2}. Both the agent and the simulated user are powered by DeepSeekV3.2~\cite{deepseekai2025deepseekv32pushingfrontieropen}. We record task success, final context length, and cumulative code execution errors. A key confound is that some tasks \textit{intrinsically} require longer interaction to solve, independent of pollution. To disentangle these effects, we stratify tasks by success rates from official benchmark evaluations~(Appendix~\ref{appendix:complexity}).

\begin{table}[htbp]
\centering
\footnotesize 
\addtolength{\tabcolsep}{-4pt}

\begin{tabular}{lcccccc}
\toprule
\multirow{2}{*}{\textbf{Difficulty}} & \multicolumn{2}{c}{\textbf{ReAct}} & \multicolumn{2}{c}{\textbf{CodeAct}} & \multirow{2}{*}{\textbf{Diff.}} \\
\cmidrule(lr){2-3} \cmidrule(lr){4-5}
 & SR & CL & SR & CL& \\
\midrule
Low & 88.2 & 7.8K & 82.4 & 12.4K & -5.8 \\
Medium & 62.5 & 10.5K & 60.0 & 16.3K & -2.5 \\
High & 19.2 & 10.7K & 26.9 & 21.1K & +7.7 \\
\midrule
Overall & 60.0 & 9.4K & 59.0 & 15.7K & -1.0 \\
\bottomrule
\end{tabular}
\caption{\footnotesize Performance comparison of ReAct vs. CodeAct methods by difficulty level. Metrics: SR (success rate per run), CL (average context token length), Diff. (CodeAct SR $-$ ReAct SR).}
\addtolength{\tabcolsep}{4pt}
\label{tab:agent_comparison}
\end{table}

\paragraph{Results}
Table~\ref{tab:agent_comparison} shows that CodeAct incurs more context length across all difficulty levels compared to ReAct. However, this overhead yields inconsistent returns: CodeAct underperforms ReAct on Low ($-$5.8\%) and Medium ($-$2.5\%) difficulty tasks, while outperforming on High difficulty tasks (+7.7\%). Figure~\ref{fig:pre_analysis} provides further insight: within the CodeAct runs, context length negatively correlates with success on both Low and High difficulty tasks, but shows no clear pattern on Medium tasks.

\paragraph{Analysis}
This pattern reveals where code-as-action provides value and where it incurs cost. Low-difficulty tasks require minimal reasoning and no iterative refinement; CodeAct's additional context is pure overhead. High-difficulty tasks benefit from code expressiveness for complex logic and multi-step operations, explaining CodeAct's +7.7\% advantage. Yet even on High tasks, the absolute success rate remains low (26.9\%) and context length is substantial (21.1K tokens). The negative correlation between context length and success within High-difficulty tasks suggests that \textit{cross-sub-task trace accumulation} still degrades performance, even when code flexibility helps. This motivates \textsc{CodeDelegator}: retain the expressiveness benefits of code-as-action for complex sub-tasks while isolating each sub-task's execution to prevent traces from polluting subsequent planning. Our design principles are detailed in Appendix~\ref{sec:design_principle}.
\section{Methodology}

\begin{figure*}
    \centering
    \resizebox{\textwidth}{!}{
    \includegraphics[scale=0.2]{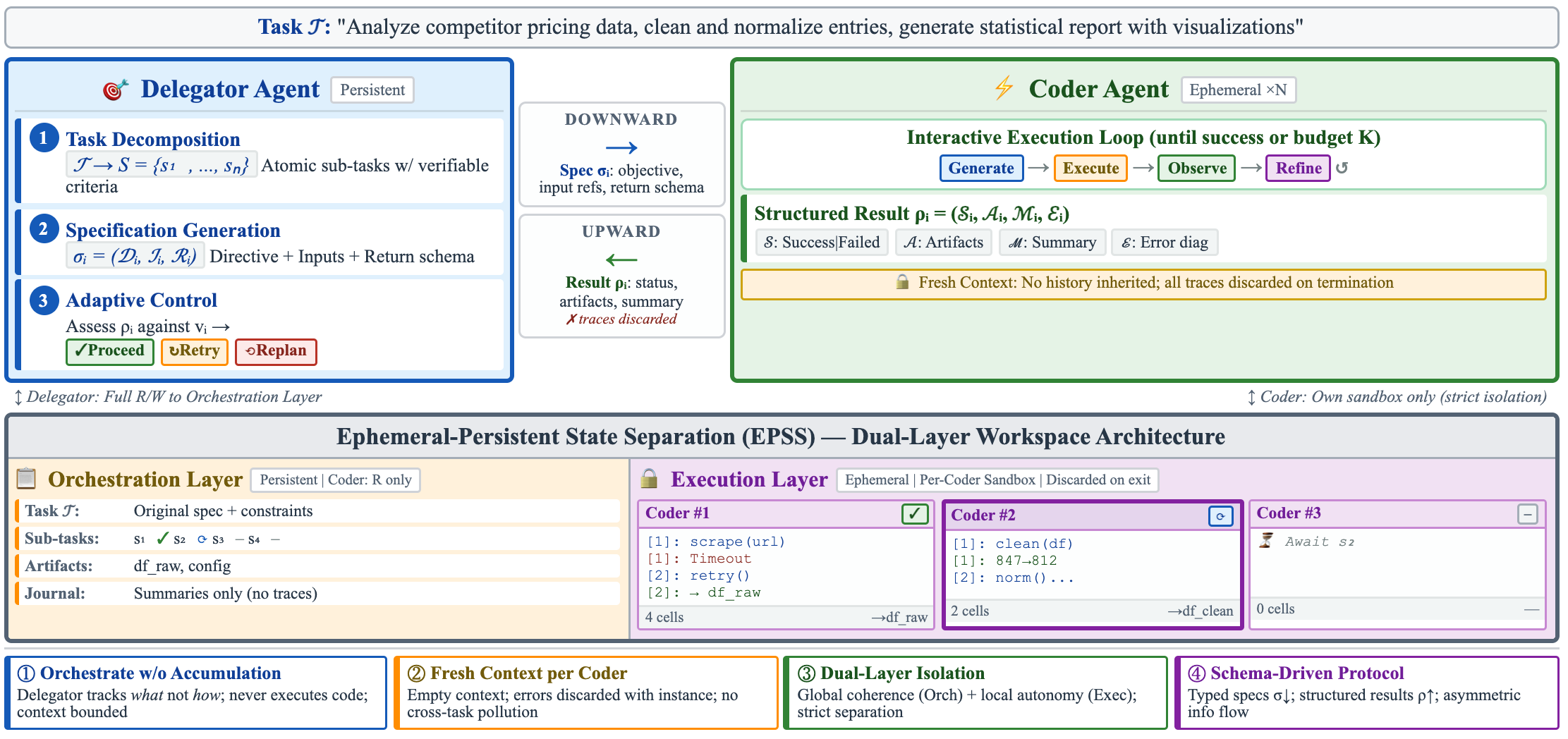}}
    \caption{Overview of the CodeDelegator framework with two core mechanisms. \textbf{Role Separation} (top): A persistent Delegator decomposes tasks and performs adaptive control, while ephemeral Coders execute sub-tasks through interactive refinement with fresh contexts. \textbf{EPSS} (bottom): The dual-layer workspace architecture isolates execution traces in per-Coder sandboxes (Execution Layer) while maintaining global coherence in a persistent Orchestration Layer, enabling four key properties: orchestration without accumulation, fresh context per Coder, dual-layer isolation, and schema-driven communication.}
\label{fig:method_overview}
\end{figure*}

\label{sec:methodology}

We introduce \textbf{\textsc{CodeDelegator}}, a multi-agent framework that separates strategic planning from code implementation through role specialization. A persistent \textit{Delegator} handles task decomposition and orchestration, while ephemeral \textit{Coders} implement atomic sub-tasks in isolated contexts. To coordinate agents with different lifespans and responsibilities, we propose \textbf{Ephemeral Persistent State Separation (EPSS)}, a dual-layer workspace where only validated outputs propagate upward. Fig.~\ref{fig:method_overview} provides an overview of the framework.

\subsection{Agent Architecture}
\label{sec:roles}

\textsc{CodeDelegator} employs two agent types with asymmetric responsibilities, lifespans, and context access patterns.

\paragraph{Delegator Agent}
\label{sec:delegator}

The Delegator is a long-lived agent that maintains strategic oversight throughout the task lifecycle. It performs three core functions:

\subparagraph{Task Decomposition}
Given a complex task $\mathcal{T}$, the Delegator decomposes it into a sequence of sub-tasks $S = \{s_1, \ldots, s_n\}$. The Delegator is instructed to identify logical sub-units that can be implemented independently, specify their execution order, and ensure each sub-task has a clear completion criterion. 
The decomposition is guided by two principles. {Atomicity}: Each sub-task should be small enough to complete within a single Coder session. {Verifiability}: Each sub-task should have a clear expected outcome that the Delegator can assess upon completion. If a sub-task later proves too complex, the Replan mechanism allows dynamic re-decomposition.
\subparagraph{Sub-task Delegation.}
\label{subpara:spec}
For each sub-task $s_i$, the Delegator constructs a structured specification $\sigma_i = (\mathcal{D}_i, \mathcal{I}_i, \mathcal{R}_i)$ comprising:
\begin{itemize}[topsep=0pt, partopsep=0pt, itemsep=1pt, parsep=0pt]
    \item $\mathcal{D}_i$: A {directive} describing the goal and constraints in natural language.
    \item $\mathcal{I}_i$: {Input bindings} as references to Orchestration Layer objects, annotated with type, shape, and sample values.
    \item $\mathcal{R}_i$: A {return schema} specifying expected output names, types, and validation conditions.
\end{itemize}
Then the Delegator delegates this sub-task specification to a new initialized Coder. The communication example is illustrated in Fig.~\ref{fig:comm_example}.

\subparagraph{Adaptive Progress Control}
After each Coder terminates, the Delegator assesses the structured result $\rho_i$~(\S~\ref{subpara:res}) to determine whether the sub-task has been satisfactorily completed. This assessment is based on the Coder's reported status, the output summary $\mathcal{M}_i$, and any error diagnostics $\mathcal{E}_i$. The Delegator then selects one of three actions:
\begin{itemize}[topsep=0pt, partopsep=0pt, itemsep=1pt, parsep=0pt]
    \item \textsc{Proceed}: Commit the successful Coder's outputs to the Orchestration workspace layer and advance to the next sub-task.
     \item \textsc{Retry}: Spawn a new Coder with a refined specification when the Delegator judges the failure as recoverable—typically specification ambiguity or missing context rather than fundamental decomposition flaws.
    \item \textsc{Replan}: Revise the overall task decomposition when the Delegator attributes failure to structural issues or when the Retry limit is reached.
\end{itemize}

\paragraph{Coder Agent}
\label{sec:coder}

Each Coder is a short-lived specialist implementing exactly one atomic sub-task. Its defining characteristic is context isolation: it inherits no history from the Delegator's planning process or sibling Coders' implementations.

\subparagraph{Interactive Execution}
A fresh Coder instance receives the structured specification $\sigma_i$, runtime access to designated input variables, and an isolated execution environment. It operates in an interactive loop analogous to human developers working with computational notebooks: (1) Generate code based on the specification and current execution state. (2) Execute code and observe outputs or errors. (3) Refine code based on execution feedback. (4) Repeat until success or budget exhaustion.
This interactive paradigm enables iterative refinement where runtime errors become actionable feedback rather than unrecoverable failures. The loop terminates when the Coder produces all outputs specified in $\mathcal{R}_i$, encounters an unrecoverable error, or exhausts its iteration budget $K$.

\subparagraph{Structured Result}
\label{subpara:res}
Upon completion, each Coder returns a structured result $\rho_i = (\mathcal{S}_i, \mathcal{A}_i, \mathcal{M}_i, \mathcal{E}_i)$ contains four components defined as follows:

\begin{itemize}[topsep=0pt, partopsep=0pt, itemsep=1pt, parsep=0pt]
    \item $\mathcal{S}_i \in \{\textsc{Success}, \textsc{Fail}\}$: The completion status. \textsc{Success} indicates all outputs satisfy the return schema. \textsc{Fail} indicates the Coder encountered issues preventing success.
    \item $\mathcal{A}_i$: Output artifacts as references to Python objects satisfying the return schema $\mathcal{R}_i$. Populated only on \textsc{Success}.
    \item $\mathcal{M}_i$: A summary of the outcome.
    \item $\mathcal{E}_i$: Brief error diagnostics with root-cause explanation. Included only on \textsc{Fail}.
\end{itemize}  
Critically, full execution histories, intermediate code cells, debugging traces and failed attempts are not returned. This information is discarded with the Coder instance, enforcing asymmetric information flow that prevents context pollution. One might worry that discarding traces loses nuance for retry decisions. We accept this trade-off deliberately: our pilot study (\S\ref{sec:analysis}) shows that trace accumulation is precisely what degrades long-horizon performance. Instead, we rely on structured diagnostics $\mathcal{E}_i$. Coders summarize root causes and failed operations, preserving what the Delegator needs to decide between \textsc{Retry} and \textsc{Replan}, while filtering implementation noise that would otherwise pollute its planning context.

\subsection{Ephemeral-Persistent State Separation}
\label{sec:workspace}

Effective multi-agent collaboration requires balancing global coherence with local autonomy. A naive approach, simple context separation, gives each agent an independent conversation history. However, code-as-action agents require deeper isolation: they exchange actual Python objects that lose fidelity when described as text, operate in interpreter namespaces where variable collisions cause subtle bugs, and generate verbose traces that must be filtered rather than merely truncated. We propose \textbf{Ephemeral-Persistent State Separation (EPSS)}, a code-aware dual-layer architecture isolating three dimensions: 
conversation context, execution namespace, and runtime artifacts (Table~\ref{tab:workspace}).

\paragraph{Orchestration Layer (Persistent)}
As illustrated in Fig~\ref{fig:method_overview} , the Delegator maintains a session-wide workspace providing task tracking, typed artifact storage and event logging:
\begin{itemize}[topsep=0pt, partopsep=0pt, itemsep=1pt, parsep=0pt]
    \item Task specification $\mathcal{T}$ and decomposition 
      $S$ with status tracking.
    \item {Committed artifacts}: Python objects from 
      successful Coders, indexed by name with type annotations.
    \item Progress journal recording outcome summaries.
\end{itemize}
Artifacts are stored as actual objects, not textual descriptions. When sub-task 2 produces \texttt{df\_clean}, subsequent Coders receive the DataFrame itself, preserving type, shape, and content without lossy natural language encoding.

\paragraph{Execution Layer (Ephemeral)}
Each Coder operates in an {isolated Python runtime}:
\begin{itemize}[topsep=0pt, partopsep=0pt, itemsep=1pt, parsep=0pt]
    \item {Namespace isolation}: Variables do not collide across Coders or persist across retries.
    \item {Trace confinement}: Stack traces, print outputs, and intermediate results remain local.
    \item {Clean state}: Deterministic environment without pollution from prior executions.
\end{itemize}
Upon termination, the runtime is discarded entirely. Only the structured result $\rho_i$, containing artifact {references}, returns to the Delegator.

\paragraph{Typed, Asymmetric Communication}
Unlike systems relying on natural language handoffs, EPSS uses schema-driven messaging:
\begin{itemize}[topsep=0pt, partopsep=0pt, itemsep=1pt, parsep=0pt]
    \item {Downward}: Specification $\sigma_i$ with typed input bindings (object references + annotations).
    \item {Upward}: Result $\rho_i$ with artifact references, summary, and diagnostics but no raw execution traces.
\end{itemize}
This enables Coders to manipulate large objects without printing them into context, while the Delegator maintains a compact $O(n)$ planning state.
\subsection{Execution Protocol}
\label{sec:protocol}

Algorithm~\ref{alg:main} formalizes the execution flow. The Delegator first decomposes task $\mathcal{T}$ into subtasks $S = \{s_1, \ldots, s_n\}$. For each subtask whose dependencies are satisfied, the Delegator generates a specification and spawns a fresh Coder, which tackles the subtask through iterative coding with tool invocations until either success or the iteration budget $K$ is exhausted. Based on the returned status and error diagnostics $\mathcal{E}_i$, the Delegator evaluates subtask completion and decides the next action. On success with evaluation criteria met, it commits artifacts to the Orchestration Layer and proceeds to the next subtask. On failure or unmet evaluation criteria, it analyzes the diagnostics to determine the cause: if the issue is resolvable through specification refinement, it retries with a fresh Coder instance; otherwise, it invokes \textsc{Replan} to split the problematic subtask, add missing dependencies, or reformulate requirements. Execution terminates successfully when all subtasks complete, or fails when retry or replan budgets are exhausted.

\section{Experiments}
\label{sec:exp}
\subsection{Experiment Setup}

\begin{table*}[t]
\footnotesize
\centering
\begin{tabular}{lllcccc}
\toprule$\textbf{Model}$ & $\textbf{Domain}$ & $\textbf{Method}$ 
& ${pass\textasciicircum 1}$ & ${pass\textasciicircum 2}$ 
& ${pass\textasciicircum 3}$ & ${pass\textasciicircum 4}$ \\
\midrule
\multirow{8}{*}{DeepseekV3.2}
    & \multirow{4}{*}{Retail}
     & ReAct               & 79.6 & 69.9 & 63.4 & \underline{58.8} \\
    & & CodeAct             & 70.2 & 59.0 & 50.0 & 47.0 \\
    & & \textsc{CodeDelegator} & \textbf{82.0} & \textbf{71.2} & \textbf{63.4} & 57.0 \\
\cmidrule(l){2-7}
    & \multirow{4}{*}{Airline}
     & ReAct               & 58.5 & 47.0 & 40.5 & 36.0 \\
    & & CodeAct             & 59.0 & 45.3 & 37.0 & 32.0 \\
    & & \textsc{CodeDelegator} & \textbf{63.5} & \textbf{53.7} & \textbf{48.5} & \textbf{46.0} \\
\midrule
\multirow{8}{*}{GPT5.0}
    & \multirow{4}{*}{Retail}
     & ReAct               & 78.7 & 68.9 & 62.9 & 58.8 \\
    & & CodeAct             & 75.0 & 65.3 & 59.3 & 57.1 \\
    & & \textsc{CodeDelegator} & \textbf{80.9} & \textbf{70.3} & \textbf{64.5} & \textbf{60.5} \\
\cmidrule(l){2-7}
    & \multirow{4}{*}{Airline}
     & ReAct               & 57.0 & 49.3 & \underline{45.0} & 42.0 \\
    & & CodeAct             & 56.5 & 47.3 & 41.5 & 40.0 \\
    & & \textsc{CodeDelegator} & \textbf{62.0} & \textbf{50.7} & 44.5 & \textbf{42.0} \\
\bottomrule
\end{tabular}
\caption{\footnotesize Results on $\tau^2$-bench. We report pass\textasciicircum k accuracy (\%) for $k \in \{1,2,3,4\}$}
\label{tab:tau}
\end{table*}

\begin{table}[t]
\footnotesize
\centering
\begin{tabular}{l ccc}
\toprule$\textbf{Domain}$ & ReAct & CodeAct & \textsc{CodeDelegator} \\
\midrule
Filesystem   & 35.0 & 39.0 & $\textbf{49.2}$ \\
GitHub       & 18.5 & 27.9 & $\textbf{38.0}$ \\
Notion       & 13.4 & 19.4 & $\textbf{34.8}$ \\
Playwright   & 13.0 & 9.7  & $\textbf{27.0}$ \\
PostgreSQL   & \underline{52.4} & 36.2 & 41.7 \\
\midrule
Total        & 25.8 & 26.4 & $\textbf{38.4}$ \\
\bottomrule
\end{tabular}
\caption{Results on MCPMark (DeepSeekV3.2). We report pass@1 (\%).}
\label{tab:mcp}
\end{table}

\paragraph{Environments}
We evaluate \textsc{CodeDelegator} under two interaction paradigms. (1) $\textbf{Human-Agent-Environment interaction}$~\textbf{(HAE)} (\S\ref{sec:exp1}): the agent converses with users while invoking tools to interact with the environment. We adopt $\tau^2$-bench~\citep{barres2025tau2}, which simulates customer service scenarios involving multi-turn dialogues and backend operations (e.g., order lookup, ticket modification), covering retail (115 tasks) and airline (50 tasks) domains. (2)~$\textbf{Agent-Environment-only interaction}$~\textbf{(AE)} (\S\ref{sec:exp2}): the user specifies a task upfront, after which the agent interacts with the environment autonomously without further dialogue. We use MCPMark~\citep{wu2025mcpmark}, which provides 127 expert-curated tasks requiring extensive CRUD operations with 5 MCP services: Filesystem, GitHub, Notion, Playwright and PostgreSQL. Together, these benchmarks assess agent capabilities across diverse interaction patterns and complexity levels.

\paragraph{Baselines}
We compare against two representative action-representation paradigms spanning the expressiveness spectrum. 
(1)~\textit{ReAct}~\cite{yao2023react}: interleaves reasoning and action with structured text, enabling explicit deliberation but limited to one tool invocation per step. (2)~\textit{CodeAct}~\cite{wang2024executable}: employs executable Python as the action space, supporting control flow and multi-tool composition within single actions. 
To ensure compatibility with contemporary tool-use frameworks and model completion APIs, we adapt CodeAct's original free-form code format to a JSON-based function call interface where code is passed as a string argument—preserving its expressiveness while enabling standardized evaluation.
Both baselines operate as single-agent systems with shared context across the entire task, isolating the effect of our proposed hierarchical delegation and EPSS mechanisms.

\paragraph{Evaluation}
For $\tau^2$-bench, we report pass\textasciicircum k (k $\in$ {1, 2, 3, 4}) (\%)~\cite{yao2024tau} across its two domains: retail and airline. A task is considered passed only if all user requirements are satisfied and all backend state changes are correct. For {MCPMark}, we report pass@1 as the primary metric, where success requires correct tool invocations, proper output parsing, and ground-truth-matching results verified by programmatic scripts. 

\paragraph{Implementation Details}
For $\tau^2$-bench, we employ DeepSeekV3.2~\cite{deepseekai2025deepseekv32pushingfrontieropen} in fast-thinking mode and GPT-5~\cite{openai2025gpt5} at low reasoning level as backbone models to ensure fair comparison, and following the $\tau^2$-bench framework requirements which require an LLM-based user simulator for multi-turn dialogue, we use DeepSeekV3.2 as the simulated user. For MCPMark, we adopt DeepSeekV3.2 as the sole backbone model due to cost considerations. The temperature is set to 0 across all experiments to ensure reproducibility. For ReAct and CodeAct baselines, we set the maximum number of interaction steps to 200 and the maximum tolerated errors to 10. For \textsc{CodeDelegator}, the Delegator component is limited to 100 dispatch rounds while each Coder instance is constrained to 20 iterations, where the Delegator and Coders share the same base model but operate within separate contexts.

\subsection{Results on HAE Scenario}
\label{sec:exp1}

Table~\ref{tab:tau} presents results on $\tau^2$-bench. With DeepSeekV3.2, \textsc{CodeDelegator} achieves the highest ${pass\textasciicircum 1}$ accuracy across both domains, reaching 82.0\% on Retail and 63.5\% on Airline. CodeAct consistently underperforms ReAct despite its greater expressiveness, confirming that context pollution negates the benefits of code-based actions in long-horizon tasks. The advantage of \textsc{CodeDelegator} becomes more pronounced under stricter consistency requirements: on Airline, the ${pass\textasciicircum 1} \to {pass\textasciicircum 4}$ degradation is 17.5\% for \textsc{CodeDelegator} versus 27.0\% for CodeAct, with \textsc{CodeDelegator} outperforming CodeAct by 14.0\% absolute at ${pass\textasciicircum 4}$ . The Airline domain, involving longer action sequences and more complex API operations, better reveals the benefits of role separation by isolating execution traces from subsequent planning steps. Results with GPT-5 show similar trends, validating that context pollution is a model-agnostic limitation.

\subsection{Results on AE Scenario}
\label{sec:exp2}

Table~\ref{tab:mcp} presents results on MCPMark across five MCP tool domains. \textsc{CodeDelegator} achieves the highest overall success rate at 38.4\%, outperforming both ReAct (25.8\%) and CodeAct (26.4\%) by approximately 12\% absolute. Despite its greater expressiveness, CodeAct performs nearly identically to ReAct, consistent with our findings that context pollution offsets the flexibility of code-based actions. The advantage of \textsc{CodeDelegator} is particularly pronounced in domains requiring complex multi-step interactions: GitHub (+10.1\% over CodeAct), Notion (+15.4\%), and Playwright (+17.3\%), where role separation and EPSS provide the greatest benefit for tracking intermediate state across API calls. The exception is PostgreSQL, where ReAct achieves the highest accuracy at 52.4\%. We attribute this to transactional semantics: SQL statements wrapped in code loops can cause partial failures that leave inconsistent states, whereas ReAct's single-call-per-step pattern naturally aligns with atomic transaction boundaries.



\section{Analysis}

\paragraph{Ablation Study}
We conduct an ablation study on MCPMark to validate the effectiveness of our proposed components, as shown in Table~\ref{tab:ablation}.
\begin{itemize}[topsep=0pt, partopsep=0pt, itemsep=1pt, parsep=0pt]
    \item $\textbf{w/o EPSS}$: Replacing EPSS and structured asymmetric communication with natural language message passing causes a 4.7\% degradation, indicating that structured protocols and isolated workspaces reduce information loss and ambiguity in inter-agent coordination.
    \item $\textbf{w/o Role Separation}$: Further replacing the dual-agent architecture with a single code-as-action agent, while retaining basic sub-task tracking and variable management from EPSS, leads to a 10.5\% drop. This confirms the benefit of role specialization in isolating planning from implementation.
\end{itemize}

\begin{table}[t]
\footnotesize
\centering
\begin{tabular}{lc}
\toprule
\textbf{Variant} & \textbf{Total} \\
\midrule
\textsc{CodeDelegator} (full) & 38.4 \\
\quad w/o EPSS & -4.7  \\
\quad w/o Role Separation & -10.5 \\
\bottomrule
\end{tabular}
\caption{\footnotesize Ablation study on MCPMark}
\label{tab:ablation}
\footnotesize
\end{table}

\begin{figure}[h!]
    \centering
    \resizebox{\columnwidth}{!}{
    \includegraphics[scale=0.25]{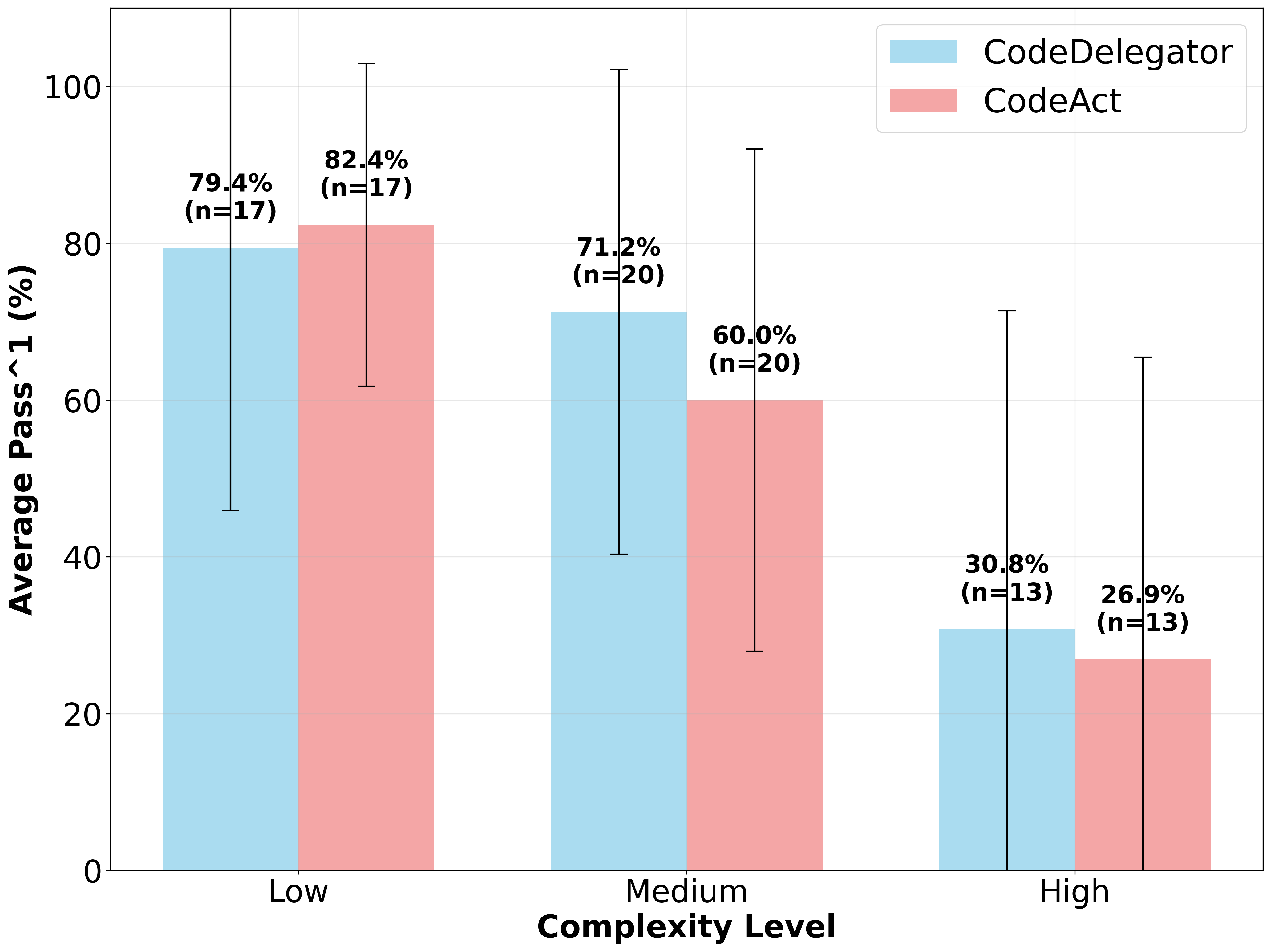}}
    \caption{$pass \textasciicircum 1$ accuracy by task complexity.}
    \label{fig:exp_analysis}
\end{figure}

\paragraph{Analysis on Complexity Scaling}
Following the task complexity stratification in Appendix~\ref{appendix:complexity}, we analyze performance across difficulty levels on $\tau^2$-bench airline domain using DeepSeekV3.2 (Fig.~\ref{fig:exp_analysis}). On Low-complexity tasks, both methods perform comparably, however, \textsc{CodeDelegator}'s advantage emerges with increasing complexity: it outperforms CodeAct by 11.2\%~(18.7\% relatively) on Medium tasks and 3.9\%~(14.5\% relatively) on High tasks. Critically, CodeAct degrades sharply from Low to High complexity, while \textsc{CodeDelegator} exhibits more graceful degradation. This pattern validates our hypothesis that role separation becomes increasingly valuable as complexity grows, with context and  runtime isolation preventing execution traces from polluting strategic planning.

\section{Conclusion}

We presented \textsc{CodeDelegator}, a multi-agent framework that addresses context pollution in code-as-action agents through role separation. A persistent Delegator handles strategic planning while ephemeral Coders implement sub-tasks in isolated contexts, preventing debugging traces from polluting the planning process. Ephemeral-Persistent State Separation (EPSS) coordinates agents through a dual-layer workspace where only validated results propagate upward. Experiments on $\tau^2$-bench and MCPMark demonstrate improvements over CodeAct, especially on hard tasks. Ablation studies confirm that role separation and EPSS contribute meaningfully to performance.

\section*{Limitations}

Our current implementation of \textsc{CodeDelegator} adopts a sequential task decomposition strategy, which simplifies planning but does not yet support more expressive agent patterns such as directed acyclic graphs (DAGs) of sub‑tasks or asynchronous interaction with the environment. In real‑world scenarios, tasks often involve parallelizable subtasks or require concurrent tool execution (e.g., monitoring data streams while processing other steps). The current linear, synchronous agent may thus underutilize resources and incur unnecessary latency. Extending \textsc{CodeDelegator} to reason about DAG‑structured plans and to support asynchronous execution and communication while preserving the isolation guarantees across multiple agents remains an important direction for future work.

\bibliography{custom}

\appendix
\section{Task Complexity in $\tau^2$-bench}
~\label{appendix:complexity} 
We analyze the trajectories from the official repository of $\tau^2$-bench (airline domain), which includes results from Claude-3.7-Sonnet~\citep{anthropic2025claude37}, GPT-4.1~\citep{openai2025gpt41}, GPT-4.1-mini~\citep{openai2025gpt41}, and o4-mini~\citep{openai2025o4mini}. Each model is evaluated over 4 independent runs, yielding 16 trials per task. We compute the average success rate across all trials and categorize tasks into three complexity levels: Low (SR $\geq$ 75\%), Medium (25\% $<$ SR $<$ 75\%), and High (SR $\leq$ 25\%). Table~\ref{tab:complexity} shows the distribution.
\begin{table}[H]
\footnotesize
\centering
\label{tab:difficulty_distribution}
\begin{tabular}{lc}
\toprule
\textbf{Difficulty Level} & \textbf{Number of Tasks}  \\
\midrule
High  & 13 \\
Medium  & 20  \\
Low & 17  \\
\midrule
Total & 50  \\
\bottomrule
\end{tabular}
\caption{Task Difficulty Distribution of $\tau^2$-bench (airline domain) (N=50)}
\end{table}

\label{tab:complexity}

\section{Design Principles}
\label{sec:design_principle}

Our pilot study (\S\ref{sec:analysis}) identifies context pollution 
as a key bottleneck: debugging traces from sub-task $s_i$ persist 
during $s_j$, diluting task-relevant information until planning 
capacity degrades. We derive four principles that directly target 
the mechanisms underlying this problem.

\paragraph{P1: Orchestrate Without Accumulation.}\mbox{}\\
\noindent\textit{Problem}: In single-agent systems, the same context must hold both strategic plans and low-level debugging traces. As traces accumulate, they compete for attention with planning-critical information.

\noindent\textit{Principle}: The Delegator tracks \textit{what} is accomplished via workspace state, never \textit{how}—it does not write or execute code.

\noindent\textit{Why it works}: By excluding execution traces entirely, the Delegator's context remains bounded to task structure ($O(n)$ summaries for $n$ sub-tasks), preserving attention for planning regardless of implementation complexity.

\paragraph{P2: Fresh Context, Focused Scope.}\mbox{}\\
\noindent\textit{Problem}: Our pilot study shows that context length negatively correlates with success on both Low and High difficulty tasks. Irrelevant history from prior sub-tasks creates noise that interferes with current decisions.

\noindent\textit{Principle}: Each Coder starts with empty context, receives only its specification and designated inputs, and terminates after returning a structured result. All traces are discarded with the instance.

\noindent\textit{Why it works}: By resetting context per sub-task, we prevent cross-sub-task pollution entirely. Each Coder operates at maximum signal-to-noise ratio for its specific task.

\paragraph{P3: Context-Runtime Decoupling.}\mbox{}\\
\noindent\textit{Problem:} Existing agentic frameworks serialize all execution outputs, including intermediate print statements and debugging traces, directly into the agent's textual context. This introduces three critical limitations: (1)~Structured data types are flattened into string representations, requiring costly re-parsing in subsequent reasoning steps; (2)~Verbose low-level details dilute the context, degrading attention over task-relevant information; (3)~The agent reasons over static textual snapshots rather than live object references, precluding interactive inspection.

\noindent\textit{Principle:} Decouple the reasoning context visible to the delegator, the sandboxed execution runtime, and the persistent state layer comprising typed objects and task trackers. Inter-module communication should use typed object references rather than printed string outputs.

\noindent\textit{Why it works:} This separation ensures that worker agents operate on first-class Python objects with preserved type signatures, execute within clean namespaces to avoid variable collisions, and return structured artifacts without lossy string encoding. The delegator observes only high-level summaries and typed references, maintaining both semantic fidelity and context efficiency.

\paragraph{P4: Schema-Driven Communication.}\mbox{}\\
\noindent\textit{Problem}: Natural language handoffs between agents risk information loss (under-specification) or pollution (over-sharing irrelevant details).

\noindent\textit{Principle}: Specifications ($\sigma_i$) and results ($\rho_i$) use structured schemas with typed fields. Downward: directive, inputs, return schema. Upward: status, artifacts, summary, error diagnostics.

\noindent\textit{Why it works}: Schemas enforce information asymmetry by design—Coders receive exactly what they need (complete), Delegators receive only what matters (filtered). This prevents both under-specification and context leakage.

\section{Examples of Asymmetric Communication}
Here shows the examples of asymmetric communication of EPSS.
\begin{figure}[H]
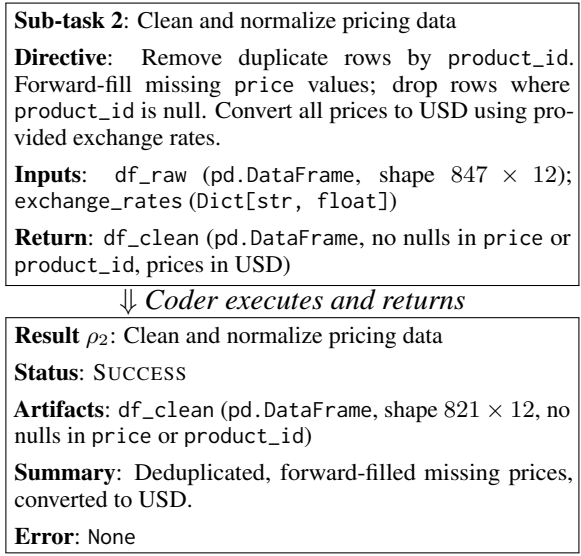

\centering
\fbox{\parbox{0.95\columnwidth}{\small$\textbf{Sub-task 2}$: Clean and normalize pricing data\\[4pt]$\textbf{Directive}$: Remove duplicate rows by \texttt{product\_id}. Forward-fill missing \texttt{price} values; drop rows where \texttt{product\_id} is null. Convert all prices to USD using provided exchange rates.\\[4pt]$\textbf{Inputs}$: \texttt{df\_raw} (\texttt{pd.DataFrame}, shape $847 \times 12$); \texttt{exchange\_rates} (\texttt{Dict[str, float]})\\[4pt]$\textbf{Return}$: \texttt{df\_clean} (\texttt{pd.DataFrame}, no nulls in \texttt{price} or \texttt{product\_id}, prices in USD)
}}

$\Downarrow$ \textit{Coder executes and returns}

\fbox{\parbox{0.95\columnwidth}{\small$\textbf{Result $\rho_2$}$: Clean and normalize pricing data\\[4pt]$\textbf{Status}$: \textsc{Success}\\[4pt]$\textbf{Artifacts}$: \texttt{df\_clean} (\texttt{pd.DataFrame}, shape $821 \times 12$, no nulls in \texttt{price} or \texttt{product\_id})\\[4pt]$\textbf{Summary}$: Deduplicated, forward-filled missing prices, converted to USD.\\[4pt]$\textbf{Error}$: \texttt{None}
}}
\caption{Asymmetric communication example. Top: specification $\sigma_i$ sent to Coder. Bottom: structured result $\rho_i$ returned to Delegator. Debugging traces remain confined to the Execution Layer.}
\label{fig:comm_example}
\end{figure}

\section{Execution Protocol Algorithm}
\begin{algorithm}[H]
\footnotesize
\caption{\textsc{CodeDelegator} Execution Protocol}
\label{alg:main}
\begin{algorithmic}[1]
\Require Task $\mathcal{T}$, retry budget $R$, iteration budget $K$
\Ensure Orchestration Layer $\mathcal{W}$ or \textsc{Failure}
\State $S \gets \textsc{Decompose}(\mathcal{T})$
\State $\mathcal{W} \gets \emptyset$
\While{$S$ has pending sub-tasks}
    \State $s_i \gets$ next pending sub-task in $S$
    \State $\textit{resolved} \gets \textsc{False}$
    \For{$r \gets 1$ to $R$}
        \State $\sigma_i \gets \textsc{GenSpec}(s_i, \mathcal{W})$
        \State $\rho_i \gets \textsc{ExecCoder}(\sigma_i, K)$
        \If{$\rho_i.\mathcal{S} = \textsc{Success}$}
            \State $\mathcal{W} \gets \mathcal{W} \cup \rho_i.\mathcal{A}$
            \State $\textit{resolved} \gets \textsc{True}$
            \State $\textbf{break}$        \ElsIf{\textsc{IsRecoverable}($\rho_i.\mathcal{E}$)}
            \State $\textbf{continue}$ \Comment{Retry with refined spec}
        \Else
            \State $S \gets \textsc{Replan}(S, s_i, \rho_i.\mathcal{E})$
            \State $\textit{resolved} \gets \textsc{True}$ \Comment{Handled via replan}
            \State $\textbf{break}$        \EndIf
    \EndFor
    \If{$\neg\textit{resolved}$}
        \State \Return \textsc{Failure}$(s_i, \rho_i.\mathcal{E})$
    \EndIf
\EndWhile
\State \Return $\mathcal{W}$
\end{algorithmic}
\end{algorithm}
 
\section{EPSS: Orchestration vs Execution Layer}
\begin{table}[H]
\setlength{\tabcolsep}{4pt}  
\centering
\resizebox{\columnwidth}{!}{
\begin{tabular}{@{}lll@{}}
\toprule$\textbf{Category}$ & $\textbf{Orch. Layer}$ & $\textbf{Exec. Layer}$ \\
\midrule
\multicolumn{3}{@{}l}{\textit{Lifecycle}} \\
Lifespan      & Session-wide         & Sub-task scoped       \\
Cardinality   & Singleton (shared)   & Per-Coder (isolated) \\
Owner         & Delegator            & Ephemeral Coder       \\
\midrule
\multicolumn{3}{@{}l}{\textit{State Management}} \\
Contents      & Sub-tasks, artifacts, journal & Code cells, local vars, traces \\
On exit       & Absorbs $\rho_i$ selectively & Entirely discarded    \\
\midrule
\multicolumn{3}{@{}l}{\textit{Information Flow}} \\
Sends ($\downarrow$) & Specs $\sigma_i$, input refs & — \\
Receives ($\uparrow$) & $\rho_i$ filtered results & — \\
Propagates ($\uparrow$) & — & Status, artifacts, summary \\
Discarded     & —                     & Debug traces, failed attempts \\
\midrule
\multicolumn{3}{@{}l}{\textit{Access Control}} \\
Delegator     & Full R/W               & None                  \\
Coder         & Input refs (R-only)    & Full R/W              \\
\bottomrule
\end{tabular}
}
\caption{Workspace architecture under EPSS.}
\label{tab:workspace}
\end{table}
\end{document}